\documentclass[letterpaper, 10 pt, conference]{ieeeconf}
\IEEEoverridecommandlockouts

\usepackage{amsmath,amssymb,amsfonts}
\usepackage{graphicx}
\usepackage{textcomp}
\usepackage{xcolor}
\usepackage{cite}
\usepackage{hyperref}
\usepackage{booktabs}
\usepackage{multirow}
\usepackage{array}
\usepackage[utf8]{inputenc}
\usepackage[T1]{fontenc}
\graphicspath{{images/}}

\begin{document}

\title{Data-Driven Optimization of Tactile Sensor Configurations for Efficient Dexterous Manipulation}

\author{Haoran Guo$^{1}$, Haoyang Wang$^{2}$, He Bai$^{3}$, \IEEEmembership{Member, IEEE}\\
Zhengxiong Li$^{4}$, \IEEEmembership{Member, IEEE}, and Lingfeng Tao$^{5}$, \IEEEmembership{Member, IEEE}%
\thanks{*This work is supported by US NSF grants \#2426269 and \#2426470.}%
\thanks{$^{1}$H. Guo is with ShanghaiTech University, School of Information Science and Technology
        (email: williamalexanda@qq.com).}%
\thanks{$^{2}$H. Wang is with University of Alberta
        (email: haoyan25@ualberta.ca).}%
\thanks{$^{3}$H. Bai is with Oklahoma State University
        (e-mail: he.bai@okstate.edu).}%
\thanks{$^{4}$Z. Li is with the University of Colorado Denver, Department of Computer Science and Engineering
        (e-mail: zhengxiong.li@ucdenver.edu).}%
\thanks{$^{5}$L. Tao is with the Department of Robotics and Mechatronics Engineering, Kennesaw State University
        (email: ltao2@kennesaw.edu).}%
}

\maketitle

% ============================================================
% ABSTRACT
% ============================================================
\noindent\begin{abstract}
Tactile sensing is critical for learning-based dexterous manipulation, yet principled guidelines for sensor placement remain largely absent. While dense sensor arrays provide rich contact feedback, they impose significant hardware costs and can even degrade policy performance by introducing redundant or conflicting inputs. This paper presents the first systematic framework for quantifying the contribution of individual tactile sensors to deep reinforcement learning (DRL) policy performance. We propose a two-stage approach: a coarse empirical pruning phase that reduces the sensor count on the Shadow Hand from 92 to 21 while retaining 93\% task performance, followed by a fine-grained active learning phase that combines Gaussian Process Regression (GPR) with Lasso regression to rank the functional importance of each remaining sensor. Our analysis reveals that sensors on the thumb, ring finger, and little finger dominate manipulation performance, while middle-finger sensors exhibit negative contributions---actively degrading policy learning. Ablation studies across three manipulation tasks (block, egg, and pen) confirm that a 14-sensor configuration preserves over 90\% of the full-array performance. Zero-shot transfer experiments on two novel objects and cross-platform validation on the Allegro and Leap Hand further demonstrate that the identified importance rankings generalize across tasks and robot morphologies. These findings establish quantitative deployment guidelines that enable practitioners to select cost-effective sensor configurations with predictable performance trade-offs.
\end{abstract}

% ============================================================
% I. INTRODUCTION
% ============================================================
\vspace{-2mm}
\section{Introduction}
\vspace{-1mm}

\noindent Tactile sensing enables robots to perceive contact, detect slip, and adapt grasp strategies in real time, making it indispensable for dexterous manipulation~\cite{ref1}. As task complexity increases, deep reinforcement learning (DRL) has emerged as a powerful framework for learning adaptive manipulation policies~\cite{ref2}, with tactile providing the most direct physical signal for capturing contact dynamics that are difficult to recover through proprioception or vision alone.

However, deploying dense tactile arrays on robotic hands introduces a fundamental tension. High-density sensor suites~\cite{ref3}---often comprising dozens to hundreds of elements distributed across phalanges and fingertips---provide rich contact information but impose substantial costs in hardware integration, calibration, and maintenance~\cite{ref4}. The resulting high-dimensional observation spaces also strain computational resources~\cite{ref5}, creating bottlenecks for real-time control on compact platforms. These challenges intensify when scaling to multi-fingered hands or full humanoid systems.

\begin{figure}[!t]
\centering
\includegraphics[width=\columnwidth]{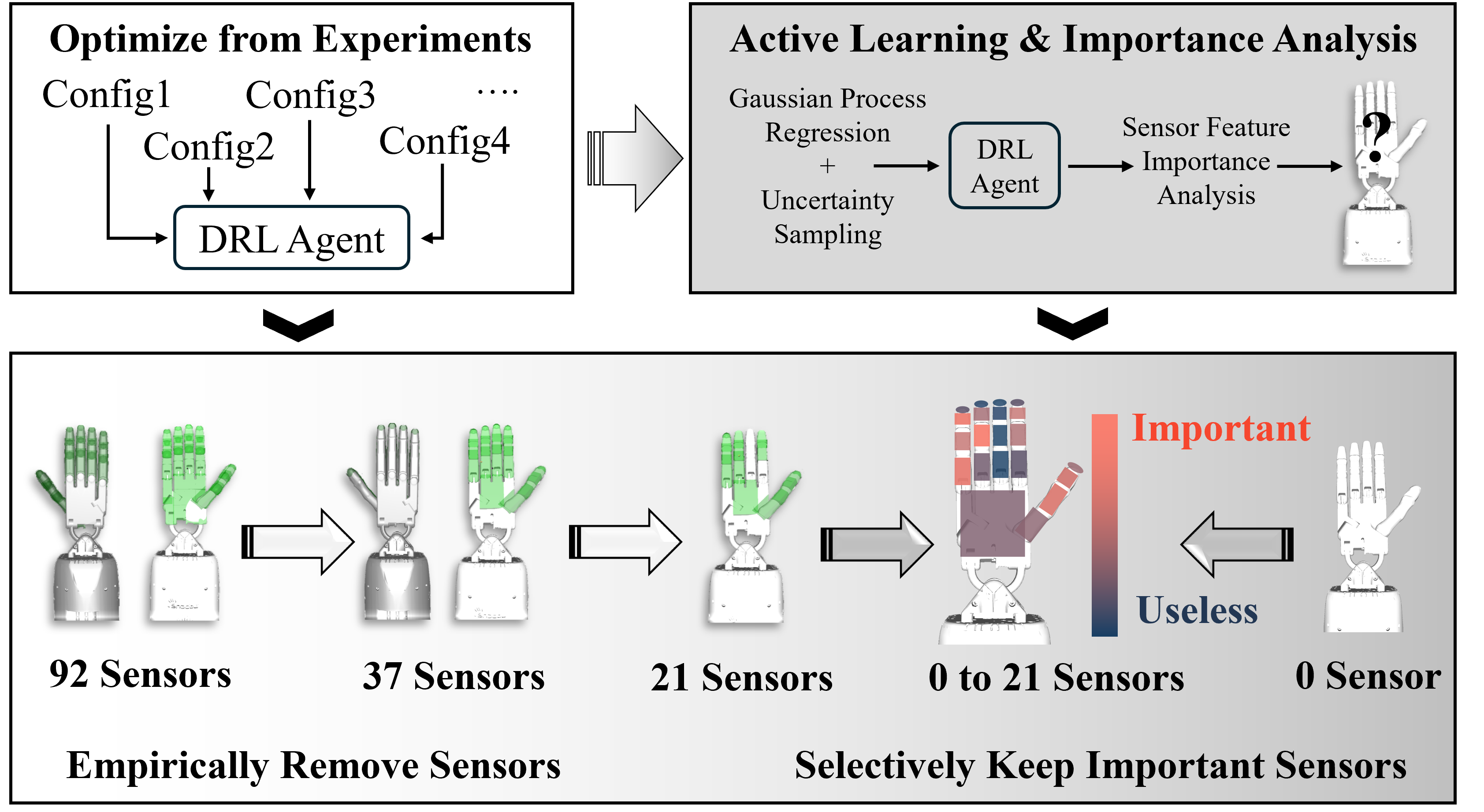}
\vspace{-6mm}
\caption{\footnotesize Overview of the tactile sensor configuration optimization framework. First, DRL policies for dexterous manipulation are trained under multiple empirical sensor configurations. By analyzing policy performance, we identify a preliminary optimized layout that reduces the sensor count from 92 to 21 while preserving task performance. Next, an active learning strategy is employed to select the most uncertain sensor configurations for additional policy training, and the resulting performance data are used to quantify the importance of individual sensors. Finally, the importance rankings are validated and leveraged to provide guidance for sensor deployment across budgets ranging from 0 to 21 sensors with best performance retention.}
\label{fig:framework}
\vspace{-6mm}
\end{figure}

A deeper problem is the absence of principled design guidelines. Unlike visual sensors~\cite{ref6}, whose placement follows well-defined field-of-view constraints, tactile sensors can be distributed across numerous candidate locations. The functional contribution of each sensor depends on the task, the learned policy, and the hand morphology in ways that are rarely intuitive. Fingertip sensors may dominate for precision grasps, while knuckle sensors capture force distributions during power grasps. When tactile data serves as input to DRL policies, this relationship becomes an opaque mapping that has seldom been systematically studied.

Recent advances in tactile simulation~\cite{ref7, ref8} create an opportunity to address this gap. High-fidelity simulators now support configurable tactile arrays, enabling rapid evaluation of diverse sensor layouts without the overhead of physical experimentation. Combined with standardized benchmark tasks~\cite{ref9}, these tools make it feasible to systematically explore the sensor configuration space.

This paper leverages this opportunity to answer a practical question: \textit{which sensors matter, and how many are actually needed?} We propose a two-stage sensor selection framework evaluated on the Shadow Hand in MuJoCo. The first stage applies coarse empirical pruning based on anatomical analysis, reducing the sensor count from 92 to 21 with minimal performance loss. The second stage employs Gaussian Process Regression (GPR) within an active learning loop to efficiently sample informative configurations from the remaining $2^{21}$ combinatorial space, followed by Lasso regression to quantify each sensor's individual contribution. The resulting importance rankings are validated across five manipulation tasks on the Shadow Hand and two additional robot platforms (Allegro Hand and Leap Hand~\cite{ref23}).

To the best of our knowledge, this is the first work to systematically quantify individual tactile sensor contributions to DRL policy performance and validate the resulting rankings across multiple tasks and robot morphologies. Our contributions are threefold:

\noindent 1) \textbf{A hierarchical sensor selection framework} that makes per-sensor importance analysis tractable on high-dimensional tactile arrays. Direct evaluation of all $2^{92}$ configurations is infeasible, and purely empirical pruning cannot quantify individual contributions. Our two-stage design---coarse anatomical pruning followed by GPR-guided active learning---bridges this gap, reducing the required evaluations to just 20 targeted experiments while providing continuous importance scores for each remaining sensor.

\noindent 2) \textbf{The discovery that tactile sensors can be actively harmful.} Contrary to the assumption that more sensors yield better performance, our analysis reveals a three-way partition: \textit{beneficial} sensors (thumb, ring, little finger) that drive performance, \textit{detrimental} sensors (middle finger) that degrade policy learning by up to 65\%, and \textit{redundant} sensors that contribute nothing. This finding---that a 14-sensor subset outperforms the 21-sensor set on certain tasks---challenges the ``dense is better'' design philosophy and provides empirical evidence that sensor pruning can serve as implicit regularization for DRL policies.

\noindent 3)  \textbf{Cross-task and cross-platform generalization.} The identified rankings hold across five manipulation tasks (including two zero-shot scenarios), three dexterous hands (Shadow, Allegro, and Leap Hand) and two physics simulation platforms, establishing them as transferable design and deployment guidelines with explicit performance-cost trade-offs.

% ============================================================
% II. RELATED WORK
% ============================================================
\vspace{-2mm}
\section{Related Work}
\vspace{-1mm}
\subsection{Tactile Sensing for Dexterous Hands}
\vspace{-1mm}
\noindent Multi-fingered dexterous hands such as the Shadow Hand have been increasingly adopted for complex manipulation tasks~\cite{ref10}. To perceive object interactions, most systems rely on either visual feedback~\cite{ref11} or tactile sensing. While vision-based methods have achieved impressive results, they are inherently limited by occlusion and cannot directly measure contact forces or material properties~\cite{ref13}.

Existing tactile designs fall into two extremes. Fingertip-centric designs~\cite{ref14} simplify integration but restrict perception to small contact regions, limiting effectiveness for tasks requiring whole-hand awareness such as in-hand reorientation~\cite{ref15}. Conversely, dense full-hand arrays~\cite{ref16} provide comprehensive coverage but introduce computational bottlenecks that reduce effective sampling rates. This tension between sensing coverage and system complexity motivates a principled method for determining which sensors are actually necessary for a given class of tasks.

\vspace{-2mm}
\subsection{Tactile Information in Learning-Based Manipulation}
\vspace{-1mm}

\noindent Tactile feedback has been shown to substantially improve DRL-based manipulation. Capacitive arrays enable material recognition through learned classifiers~\cite{ref17}, and TacGNN~\cite{ref18} demonstrates that graph neural networks can extract tactile features sufficient for vision-free in-hand manipulation. Church et al.~\cite{ref19} combined tactile data with reinforcement learning for continuous adaptive control, while Melnik et al.~\cite{ref8} showed that tactile input significantly improves sample efficiency in simulated manipulation tasks.

The most closely related work is that of Melnik et al.~\cite{ref7}, who reorganized 92 tactile sensors into 16 groups and showed that grouped configurations achieve comparable performance to ungrouped ones. However, their approach preserves the total sensor count and does not quantify the contribution of individual sensors---leaving the fundamental question of which sensors can be eliminated unanswered. Our work directly addresses this gap by providing per-sensor importance and validating it across tasks and platforms.

% ============================================================
% III. PROBLEM SETUP
% ============================================================
\vspace{-2mm}
\section{Preliminaries }
\vspace{-1mm}
\subsection{Task Formulation}
\vspace{-1mm}
\noindent We formulate the manipulation task as a Markov Decision Process $(\mathcal{S}, \mathcal{A}, \mathcal{T}, R, \gamma)$, where $\mathcal{S}$ includes tactile activations and joint angles, $\mathcal{A}$ corresponds to joint-level control commands, $\mathcal{T}$ is the transition function, $R$ is a binary reward indicating whether the object reaches the target pose, and $\gamma \in [0,1]$ is the discount factor. We adopt the Deep Deterministic Policy Gradient (DDPG)~\cite{ref20}, which learns a deterministic policy $\mu_\theta(s)$ and a critic $Q_\varphi(s,a)$ via the Bellman objective.

\vspace{-2mm}
\subsection{Sensor Layout and Categorization}
\vspace{-1mm}
\noindent Preliminary experiments use the MuJoCo~\cite{ref9} with the OpenAI Gym Shadow Hand environment, which originally incorporates 92 tactile sensors. Policy training uses hindsight experience replay (HER)~\cite{ref24} to accelerate the training. Following prior findings demonstrating comparable DRL performance between analog and binary tactile signals~\cite{ref8}, we adopt binary sensor outputs (1 for active contact, 0 otherwise). The 92 sensors are categorized by anatomical location as illustrated in Fig.~\ref{fig:sensor_categorization}:

\begin{figure}[!b]
\vspace{-6mm}
\centering
\includegraphics[width=0.7\columnwidth]{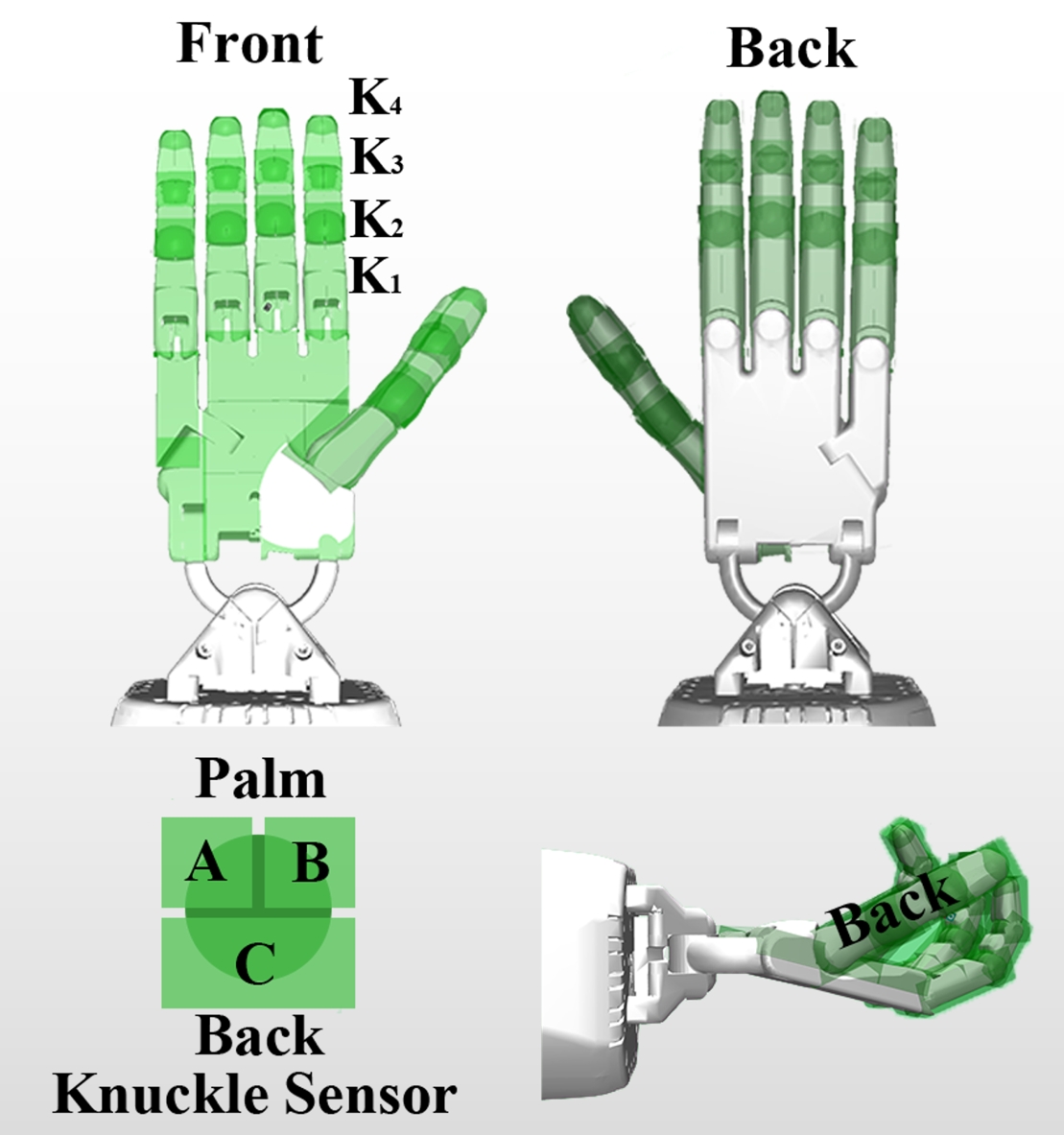}
\vspace{-2mm}
\caption{\footnotesize Sensor placement for the original Shadow Hand. $K_1$--$K_4$ denote the proximal, middle, top, and tip knuckles of each finger, respectively. Each knuckle is equipped with three sensors labeled A, B, and C.}
\label{fig:sensor_categorization}
\end{figure}

\textit{Joints} (15 sensors): Three spherical sensors per finger---one at the fingertip and two at finger joints. \textit{Knuckles--Front} (38 sensors): Four rectangular sensors per knuckle ($K_1$--$K_4$, denoting proximal to tip), capturing palm-facing and dorsal contact. \textit{Knuckles--Back} (30 sensors): Rectangular sensors on the dorsal side. \textit{Palm} (9 sensors): Sensors distributed across the palm surface and metacarpal region.

\vspace{-2mm}
\subsection{Manipulation Tasks and Training Protocol}
\vspace{-1mm}

\begin{figure}[!b]
\centering
\vspace{-2mm}
\includegraphics[width=0.8\columnwidth]{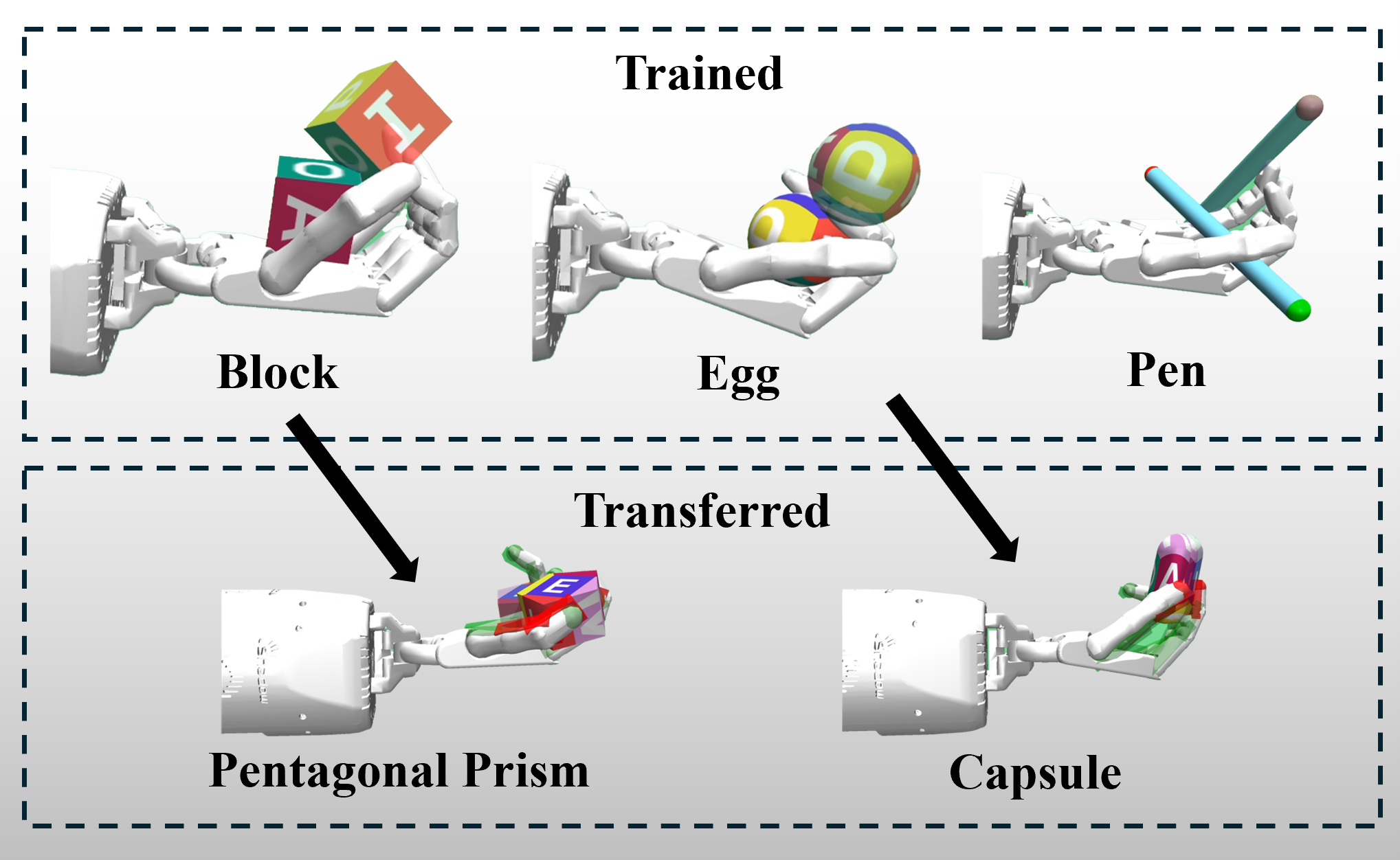}
\vspace{-2mm}
\caption{\footnotesize All manipulation tasks considered in this paper. Block, Egg, and Pen are built-in tasks of the original Gym environment; Pentagonal Prism and Capsule manipulation tasks are self-defined tasks. The Block task is used in the sensor quantity and importance analysis study, the Egg and Pen tasks are used to verify the proposed important sensor locations, and the Pentagonal Prism and Capsule tasks are used in zero-shot experiments to test the generalization ability of the proposed sensor configurations.}
\label{fig:tasks}
\end{figure}

\noindent We evaluate five manipulation tasks (Fig.~\ref{fig:tasks}): Block, Egg, and Pen as built-in Gym scenarios, plus custom Pentagonal Prism (PP) and Capsule tasks for transfer evaluation. All models are trained using 12 parallel threads for 200 epochs of 100 cycles each to ensure performance convergence. Each experiment is repeated with 3 random seeds, and we report the average success rate. To enable cross-configuration comparison, we define normalized performance:
\begin{equation}
P_{\text{Config}(i)} = \frac{S_{\text{config}(i)}}{S_{\text{Max}}} \times 100\%
\label{eq:performance}
\end{equation}
where $S_{\text{config}(i)}$ is the success rate of configuration $i$ and $S_{\text{Max}}$ is the max success rate across all configurations for that task.

Note that the absolute success rates (0.3--0.5 for the Block task under DDPG) reflect the inherent difficulty of in-hand manipulation rather than insufficient training. Our analysis targets \textit{relative} performance differences across configurations, for which this range provides sufficient sensitivity.

% ============================================================
% IV. METHODOLOGY
% ============================================================
\vspace{-2mm}
\section{Methodology}
\vspace{-1mm}

\noindent Our framework follows a hierarchical two-stage design (Fig.~\ref{fig:framework}). The first stage applies coarse pruning based on physical reasoning to rapidly eliminate sensor groups with minimal task relevance, reducing the search space from 92 to 21 sensors. The second stage performs fine-grained analysis on the remaining sensors using active learning to efficiently explore the $2^{21}$ combinatorial space and Lasso regression to quantify individual sensor importance. This hierarchical approach balances computational efficiency with analytical precision---direct application of active learning to all 92 sensors would be intractable, while purely empirical pruning cannot quantify per-sensor contributions.

\vspace{-2mm}
\subsection{Stage 1: Coarse Empirical Pruning}
\vspace{-1mm}
\noindent We systematically reduce the sensor array through a sequence of physically motivated simplifications, where each step targets a specific source of redundancy identified from contact mechanics and observed sensor activation patterns. Each configuration $A_q$ ($q$ = number of sensors) is evaluated under identical training conditions:

\begin{itemize}
    \item $A_{92}$: Full sensor set (baseline).
    \item $A_{62}$: Dorsal sensors removed ($-30$). During in-hand manipulation, objects primarily contact the palmar surface; dorsal sensors showed near-zero activation rates across training episodes, contributing no informative gradient signal to the policy.
    \item $A_{39}$: Adjacent palmar knuckle sensors merged ($-23$). Physically co-located sensors on the same knuckle face exhibit co-activation rates exceeding 90\%, making their signals highly redundant. Merging them preserves spatial coverage while halving the knuckle input dimensionality.
    \item $A_{29}$: Finger joint sensors removed except fingertips ($-10$). Joint-level spherical sensors sit between phalanges where object contact is geometrically unlikely during stable grasps; their activation rates during converged policies remained below 5\%.
    \item $A_{21}$: Palm and metacarpal sensors consolidated ($-8$). Multiple palm sensors detect the same event---initial object contact with the palm---and can be replaced by a single unit without loss of task-relevant information.
    \item $A_{0}$: All tactile sensors removed (proprioception-only baseline).
\end{itemize}

As be elaborated in detail in the Results chapter, this pruning reduces the sensor count by 77\% (from 92 to 21) while preserving 93.1\% of the normalized performance, thereby establishing $A_{21}$ as the baseline for subsequent fine-grained analysis. Importantly, performance does not decrease monotonically with sensor count: $A_{21}$ slightly outperforms $A_{29}$, foreshadowing the finding (Section~\ref{sec:importance}) that certain sensors are not merely redundant but actively detrimental.

\vspace{-2mm}
\subsection{Stage 2: Active Learning for Sensor Selection}
\vspace{-1mm}

\noindent Exhaustively evaluating all $2^{21} = 2{,}097{,}152$ subsets of $A_{21}$ is computationally infeasible (each training run requires $\sim$24 hours). We therefore employ an active selection strategy combining targeted empirical evaluation with GPR-guided uncertainty sampling~\cite{ref21}. All sensor configurations mentioned in this section are shown in fig.5a.

\textbf{Initial data collection.} The quality of the GPR surrogate depends critically on the diversity of the initial training set. We design ten physically motivated configurations $B_i$ ($i = 1, \ldots, 10$) spanning two complementary axes of variation: \textit{intra-finger sets} that activate all sensors on a single finger (5 configurations, one per finger), and \textit{cross-finger sets} that group sensors at the same anatomical position across all five fingers (e.g., all $K_2$ knuckle sensors, all fingertip sensors). This design ensures that the data covers both finger-level and position-level effects, enabling the GPR to distinguish whether performance depends more on \textit{which finger} is instrumented or \textit{where on the finger} sensors are placed.

\textbf{Surrogate modeling.} A GPR model is trained on the evaluated configurations to predict performance across the untested space. The composite kernel combines a Radial Basis Function (RBF) with a noise term:
\begin{equation}
k(x, x') = \sigma_f^2 \exp\left( -\frac{\|x - x'\|^2}{2l^2} \right) + \sigma_n^2 \delta(x, x')
\label{eq:kernel}
\end{equation}
where $l$ controls smoothness, $\sigma_f^2$ scales signal magnitude, and $\sigma_n^2$ models observation noise. Hyperparameters are optimized via log-marginal likelihood maximization.

\textbf{Uncertainty-driven sampling.} The GPR posterior variance identifies configurations where the model is least certain. At each iteration, we select:
\begin{equation}
z_{\text{next}} = \arg\max_{z \in \mathcal{S}} \left[ k(z, z) - \mathbf{k}(z, Z)(K + \sigma_n^2 I)^{-1} \mathbf{k}(Z, z) \right]
\label{eq:active_selection}
\end{equation}
The selected configuration $C_i$ is evaluated, added to the training set, and the GPR is retrained. This loop continues until the maximum predictive standard deviation falls below 0.05. We chose this threshold because it corresponds to approximately half the performance difference between $A_{21}$ and $A_0$ divided by the number of sensors---below this level, the uncertainty in predicting any untested configuration is smaller than the expected per-sensor contribution, meaning additional evaluations would yield diminishing returns for the subsequent importance analysis.

\subsection{Sensor Importance Quantification}
\vspace{-1mm}

\noindent To extract per-sensor importance from the collected data (configurations $A_{21}$, $A_0$, $B_1$--$B_{10}$, $C_1$--$C_{10}$), we apply Lasso regression~\cite{ref22}:
\begin{equation}
\hat{\beta} = \arg\min_{\beta} \left\{ \frac{1}{2n}\sum_{i=1}^{n}(y_i-\beta_0-\mathbf{x}_i^{\top}\beta)^2 + \lambda\|\beta\|_1 \right\}
\label{eq:lasso}
\end{equation}
where $x_{ij} \in \{0, 1\}$ indicates whether sensor $j$ is active in configuration $i$, $y_i$ is the corresponding task performance, and $\lambda = 0.001$ controls sparsity. The $\ell_1$ penalty drives negligible coefficients to exactly zero, producing a sparse importance vector $\hat{\beta} = (\beta_1, \ldots, \beta_{21})$ that directly ranks sensors. Optimization uses coordinate descent with 1000 maximum iterations and convergence tolerance $10^{-4}$. The reason to choose LASSO regression is it can automatically select the most representative sensors from a redundant set. Sensors whose coefficients are driven to zero are interpreted as having negligible marginal contribution to task performance given the presence of others. By constraining the sum of absolute coefficients, LASSO reduces model complexity and mitigates overfitting to environment-specific noise, thereby improving robustness across different scenarios. 

The resulting coefficient vector partitions the 21 sensors into three categories: \textit{beneficial} ($\beta_j > 0$), whose presence improves task performance; \textit{detrimental} ($\beta_j < 0$), whose presence actively degrades performance; and \textit{redundant} ($\beta_j = 0$), which contribute no information beyond what other sensors already provide.

\subsection{Statistical Validation}
\vspace{-1mm}

\noindent To validate the significance of performance differences, we conduct paired $t$-tests across all five tasks. For each configuration, success rates are collected from 50 evaluation epochs post-convergence across three random seeds. Significance thresholds follow standard conventions: $p^{*} < 0.05$, $p^{**} < 0.01$, $p^{***} < 0.001$.

% ============================================================
% V. EXPERIMENT DESIGN
% ============================================================

\section{Experiment Design}
\vspace{-1mm}
\subsection{Sensor Importance Verification}
\vspace{-1mm}

\noindent We construct ablation groups based on the Lasso importance coefficients to verify the rankings:

\begin{itemize}
    \item $D_1$: All sensors with positive coefficients (beneficial sensors only).
    \item $D_2$: Sensors with coefficients $> 0.001$.
    \item $D_3$: Sensors with coefficients $> 0.005$.
    \item $D_4$--$D_5$: Progressively removing the lowest-ranked sensor from the preceding group until only five sensors remain.
    \item $D_{\text{last}}$ ($D_6$): All sensors with non-positive coefficients (the ``harmful/useless'' set).
\end{itemize}

Each configuration is evaluated on the Block, Egg, and Pen tasks. The key hypothesis is that performance should decrease monotonically from $D_1$ to $D_5$, and that $D_6$ should perform near or below the sensor-free baseline $A_0$.

\subsection{Cross-Task Generalization}
\vspace{-1mm}

\noindent We deploy DRL policies trained on Block and Egg directly to two novel environments---Pentagonal Prism and Capsule---without training (zero-shot transfer). Each policy is evaluated over 1{,}000 trials with results normalization. Baselines $A_{92}$, $A_{21}$, and $A_0$ are transferred under identical conditions.

\subsection{Cross-Robot Generalization}
\vspace{-1mm}

\noindent To test whether importance rankings transfer across morphologies, we evaluate four configurations ($A_{21}$, $A_0$, $D_1$, $D_6$) on the Allegro Hand and Leap Hand~\cite{ref23} using a cube rotation task in IsaacGym. Since both are four-fingered platforms, we map their finger topology to the Shadow Hand: the third finger from the thumb is designated as the middle finger, and the fourth finger assumes the combined role of ring and little fingers, inheriting the higher importance value of the two. Tactile sensors use binary outputs, and performance is normalized against $A_{21}$.

\begin{figure*}[!t]
\centering
\includegraphics[width=\textwidth]{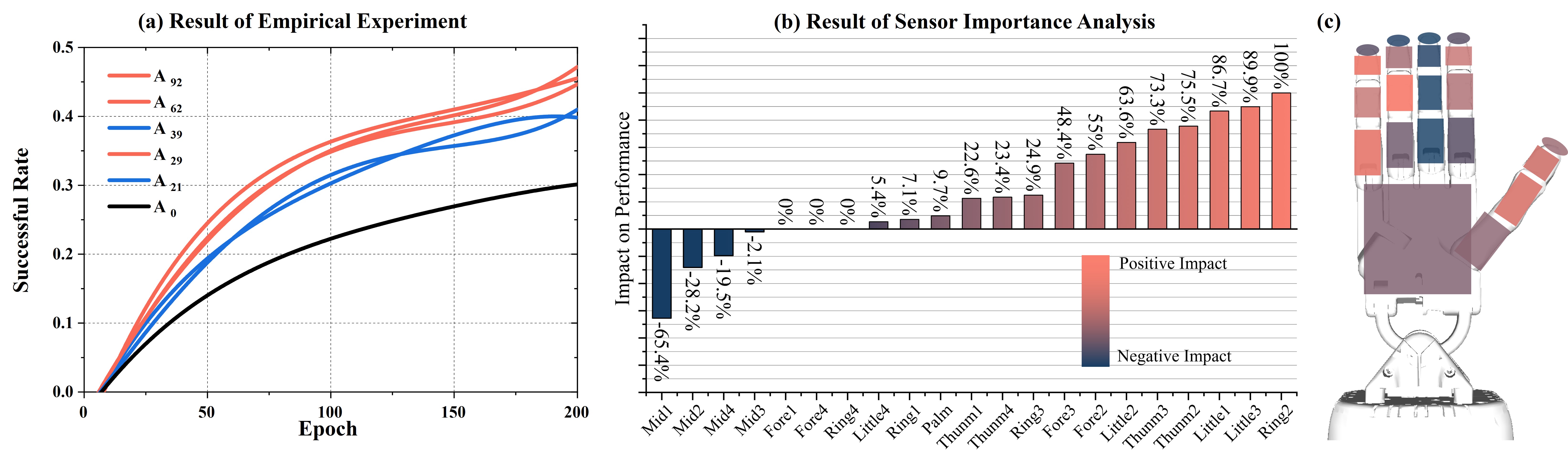}
\vspace{-2mm}
\caption{\footnotesize (a) Success rate results of empirical experiments. Orange represents a high-level, blue represents a medium-level, and black represents a low-level success rate. (b) Normalized importance of each sensor in the $A_{21}$ configuration. Sensors with different importance are marked with different colors: red represents positive impact, blue represents negative impact. (c) Location of all 21 sensors, color-coded according to their impact.}
\vspace{-6mm}
\label{fig:learning_curves}
\end{figure*}

% ============================================================
% VI. RESULTS AND DISCUSSION
% ============================================================
\vspace{-2mm}
\section{Results and Discussion}
\vspace{-1mm}

\subsection{Coarse Pruning Results}
\vspace{-1mm}

\noindent Fig.~\ref{fig:learning_curves}a shows learning curves for six configurations. All configurations improve monotonically during training, but performance does not scale linearly with sensor count. Configurations $A_{62}$ and $A_{39}$ achieve approximately 86\% of the $A_{92}$ baseline, while $A_{21}$---with only 23\% of the original sensors---retains 93.1\% performance. Given that commercial tactile sensors (e.g., uSkin) cost approximately \$5{,}000 per hand, this 77\% reduction represents substantial savings.

A non-obvious result is that $A_{29}$ (29 sensors) slightly underperforms $A_{21}$ (21 sensors). This reversal challenges the intuitive assumption that more sensors always help and provides early evidence for a more nuanced relationship between observation dimensionality and policy quality. We hypothesize that the eight additional sensors in $A_{29}$---predominantly palm and metacarpal units---provide spatially redundant signals that increase the effective dimensionality of the observation space without adding discriminative information. In the DDPG framework, this excess dimensionality can slow critic convergence by distributing gradient signal across correlated input features, an effect well-documented in the DRL literature on observation space design. This phenomenon motivates the fine-grained analysis in Stage~2, which aims to identify not only \textit{how many} sensors are needed but \textit{which specific} sensors contribute positively.

\subsection{Active Learning Efficiency}

\begin{figure*}[!t]
\centering
\includegraphics[width=\textwidth]{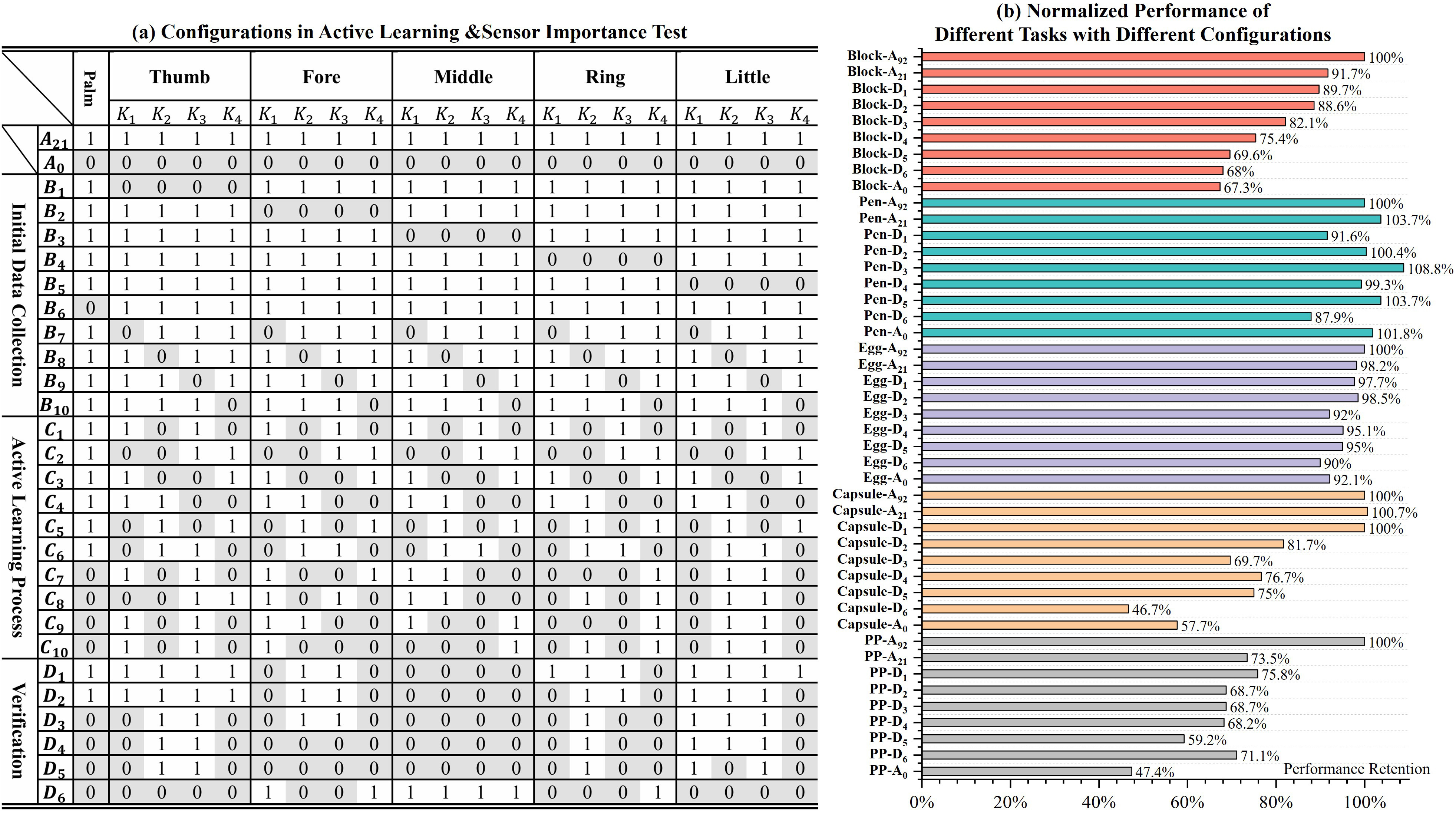}
\caption{\footnotesize (a) Detailed sensor configurations used in empirical selection ($B_i$), active learning ($C_i$), and sensor importance verification \& generalization ($D_i$). `1' on a white background indicates the presence of a tactile sensor, while `0' on gray indicates its absence. (b) Performance benchmark of Block, Egg, Pen, Pentagonal Prism (PP), and Capsule tasks. For all tasks, performance is normalized to the $A_{92}$ configuration, which serves as the 100\% baseline.}
\vspace{-4mm}
\label{fig:active_learning}
\end{figure*}

\noindent Fig.~\ref{fig:active_learning}a details the configurations sampled during active learning. The initial empirical set ($B_1$--$B_{10}$) yields a GPR model with mean predictive standard deviation of 0.0392. After evaluating the ten uncertainty-maximizing configurations ($C_1$--$C_{10}$), the retrained model achieves a standard deviation of 0.0302---a 23.0\% reduction. This demonstrates that targeted sampling efficiently maps the configuration landscape with only 20 total evaluations out of over two million possible combinations.

\subsection{Sensor Importance Rankings}
\label{sec:importance}

\noindent The Lasso regression coefficients (Fig.~\ref{fig:learning_curves}b--c) reveal a structured importance landscape with clear anatomical patterns.

\textbf{Positive-impact sensors.} The highest-ranked sensors (Ring2: $+100\%$, Little3: $+89.9\%$, Little1: $+86.7\%$, Thumb2: $+75.3\%$) concentrate on the second and third knuckles ($K_2$--$K_3$) of the thumb, ring, and little fingers. Biomechanically, these knuckle positions correspond to the intermediate phalanges, which form the primary contact surface during power-grasp phases of in-hand reorientation. The forefinger sensors (Fore2: $+73.3\%$, Fore3: $+63.5\%$) rank slightly lower, consistent with the forefinger as a guide rather than a primary force-bearing digit in the Shadow Hand's manipulation strategy.

\textbf{Negative-impact sensors.} All four middle-finger sensors receive negative coefficients, with Mid1 exhibiting the largest magnitude ($-65.4\%$), followed by Mid2 ($-28.2\%$), Mid4 ($-19.5\%$), and Mid3 ($-2.1\%$). This consistent negative pattern across an entire finger is striking and unlikely to be a statistical artifact. We attribute this to the kinematic role of the middle finger during in-hand manipulation: it is the longest digit and sits centrally between the opposing thumb and the lateral fingers, making it geometrically constrained during object rotation. As a result, the middle finger primarily serves a passive structural support function, and its sensors detect incidental contact events that are poorly correlated with task-relevant state transitions. Including these signals in the observation vector forces the policy network to learn to ignore them---an implicit feature selection burden that slows convergence and can trap the optimizer in suboptimal regions of parameter space.

\textbf{Near-zero sensors.} The palm sensor and several proximal knuckle sensors ($K_1$) receive coefficients near zero, indicating that their contributions are subsumed by neighboring sensors. These positions detect initial object contact that is already captured by the $K_2$ sensors higher on the same finger.

This finding has a practical implication: \textit{not all sensors are neutral when removed---some should be actively excluded to improve performance.} The three-way partition into beneficial, detrimental, and redundant sensors provides a more actionable framework than simple importance ranking, as it distinguishes sensors that can be safely removed (redundant) from those that \textit{should} be removed (detrimental).

\subsection{Ablation Verification}

\noindent Fig.~\ref{fig:active_learning}b benchmarks the ablation configurations against block, egg and pan tasks. The results confirm the importance rankings across tasks:

\textbf{Block task.} Performance decreases monotonically from $D_1$ (89.7\%) through $D_5$ (69.6\%), validating the ranking order. $D_6$ (68.0\%) performs comparably to $A_0$ (67.3\%), confirming that the ``harmful'' sensor set provides no useful information.

\textbf{Pen task.} $D_3$ achieves 108.8\% of the $A_{92}$ baseline, \textit{exceeding} full-array performance. Rather than an anomaly, this result reinforces our central finding that redundant sensors can actively hurt performance. The Pen task's elongated geometry concentrates effective contact onto a narrow band of sensors along the finger pads; sensors outside this band detect only transient or incidental contact that is uncorrelated with task progress. Including these uninformative dimensions increases the effective rank of the observation covariance matrix without increasing the signal-to-noise ratio, forcing the critic network to fit noise patterns that do not generalize across episodes. The effect is amplified by the Pen task's inherently low success rate ($<30\%$), which reduces the ratio of informative to uninformative training transitions. This observation provides direct empirical evidence that sensor pruning can serve as a form of \textit{implicit regularization} for DRL policies, complementing the extensive body of work on explicit regularization techniques.

\textbf{Egg task.} All configurations from $D_1$ to $D_5$ achieve 92--98.5\% performance, with $D_6$ dropping to 90\%. The compressed performance range reflects the Egg task's high baseline ($A_0 > 70\%$ success rate), creating a ceiling effect that reduces sensitivity to sensor configuration.

\textbf{Summary.} The Block task, with moderate difficulty and strong sensitivity to sensor input, serves as the most discriminative benchmark. The Pen and Egg tasks exhibit compressed performance gradients for opposite reasons---excessive difficulty and excessive ease, respectively---but the monotonic ranking trend from $D_1$ to $D_5$ holds across all three tasks.

\subsection{Cross-Task Generalization}

\noindent Zero-shot transfer results (Fig.~\ref{fig:active_learning}b, bottom part) reveal an important asymmetry between training and transfer conditions. The $D_1$ configuration achieves performance comparable to $A_{21}$ on both Pentagonal Prism (75.8\% vs.\ 73.5\%) and Capsule (100\% vs.\ 100.7\%) tasks, whereas on the training tasks, even $D_2$ (12 sensors) sufficed to match $A_{21}$. This gap---two additional sensors required for transfer parity---suggests that zero-shot generalization imposes a higher \textit{tactile information threshold} than within-distribution evaluation. Intuitively, when the policy encounters unfamiliar object geometries, it cannot rely on learned contact priors and must extract task-relevant signals from a broader sensor set. The sensors in $D_1 \setminus D_2$ (those ranked 13th--14th in importance) provide this additional coverage, serving as ``insurance'' channels that become critical only under distribution shift.

\begin{table}[!t]
\caption{Statistical Significance Analysis Results ($p$-value)}
\label{tab:significance}
\centering
\begin{tabular}{lccc}
\toprule
\textbf{Task} & $A_{92}$ vs $A_{21}$ & $A_{21}$ vs $D_1$ & $D_1$ vs $D_6$ \\
\midrule
Block   & $p^{***} < 0.001$ & n.s.  & $p^{***} < 0.001$ \\
Pen     & n.s.              & n.s.  & $p^{***} < 0.001$ \\
Egg     & n.s.              & n.s.  & $p^{***} < 0.001$ \\
\midrule
Capsule & n.s.              & n.s.  & $p^{***} < 0.001$ \\
PP      & $p^{***} < 0.001$ & n.s.  & $p^{*}\;\; < 0.05$ \\
\bottomrule
\end{tabular}
\vspace{-4mm}
\end{table}

The $D_6$ configuration produces markedly different effects across transfer tasks, providing further insight into the nature of detrimental sensors. In the Capsule task, $D_6$ achieves only 46.7\%---below even the sensor-free $A_0$ (57.7\%). This means that equipping the hand with exclusively harmful sensors is \textit{worse than having no sensors at all}: the detrimental signals actively mislead the policy, producing a negative information value that exceeds the benefit of any contact detection. In the Pentagonal Prism task, $D_6$ (71.1\%) unexpectedly surpasses $D_2$--$D_5$ ($\approx$68\%), likely because the pentagonal prism's five-fold symmetry creates a contact distribution that happens to correlate with the middle-finger sensor locations dominant in $D_6$. This geometry-specific exception does not invalidate the general ranking---$D_6$ still underperforms $D_1$ in every task---but it highlights that sensor importance has a task-dependent component that grows stronger as configurations become sparser.

Excluding this special case, all $D_1$--$D_5$ configurations maintain at least 68\% performance on the transfer tasks, consistently outperforming the sensor-free baseline. In the zero-shot experiments, the sensor-equipped groups $D_{1}$--$D_{5}$ significantly outperform the sensor-free group $A_{0}$, indicating that direct transfer of an unadapted model relies more heavily on critical tactile sensors. The presence of several key tactile sensors substantially improves the generalization capability of the learned policy.

\subsection{Statistical Significance}

\noindent Table~\ref{tab:significance} provides paired $t$-test results. The full array $A_{92}$ significantly outperforms $A_{21}$ only on the Block ($p^{***}$) and PP ($p^{***}$) tasks; no significant difference is observed on Pen, Egg, or Capsule. This asymmetry is explained by task difficulty: the Block and PP tasks occupy the moderate-difficulty regime where sensor configuration most strongly differentiates performance, while the Pen task (too hard, success rate $<30\%$) and Egg task (too easy, $A_0 > 70\%$) exhibit performance gradient compression that masks configuration effects in statistical tests.

Critically, $D_1$ maintains statistical parity (n.s.) with $A_{21}$ across \textit{all five tasks}, including zero-shot scenarios, despite containing 33\% fewer sensors. This uniform non-significance is the strongest evidence that the Lasso-identified beneficial sensor set captures the full task-relevant information content of $A_{21}$---the removed sensors contribute only noise. Meanwhile, $D_1$ significantly outperforms $D_6$ in every case ($p^{***}$ on four tasks, $p^{*}$ on PP), confirming the efficacy of the importance-based selection. The weaker significance on PP ($p^{*}$ vs.\ $p^{***}$) is consistent with the $D_6$ anomaly discussed in Section~VI-E, where the Pentagonal Prism's geometry partially compensates for the detrimental sensor set.

\subsection{Cross-Robot Generalization}

\begin{figure}[!t]
\centering
\includegraphics[width=\columnwidth]{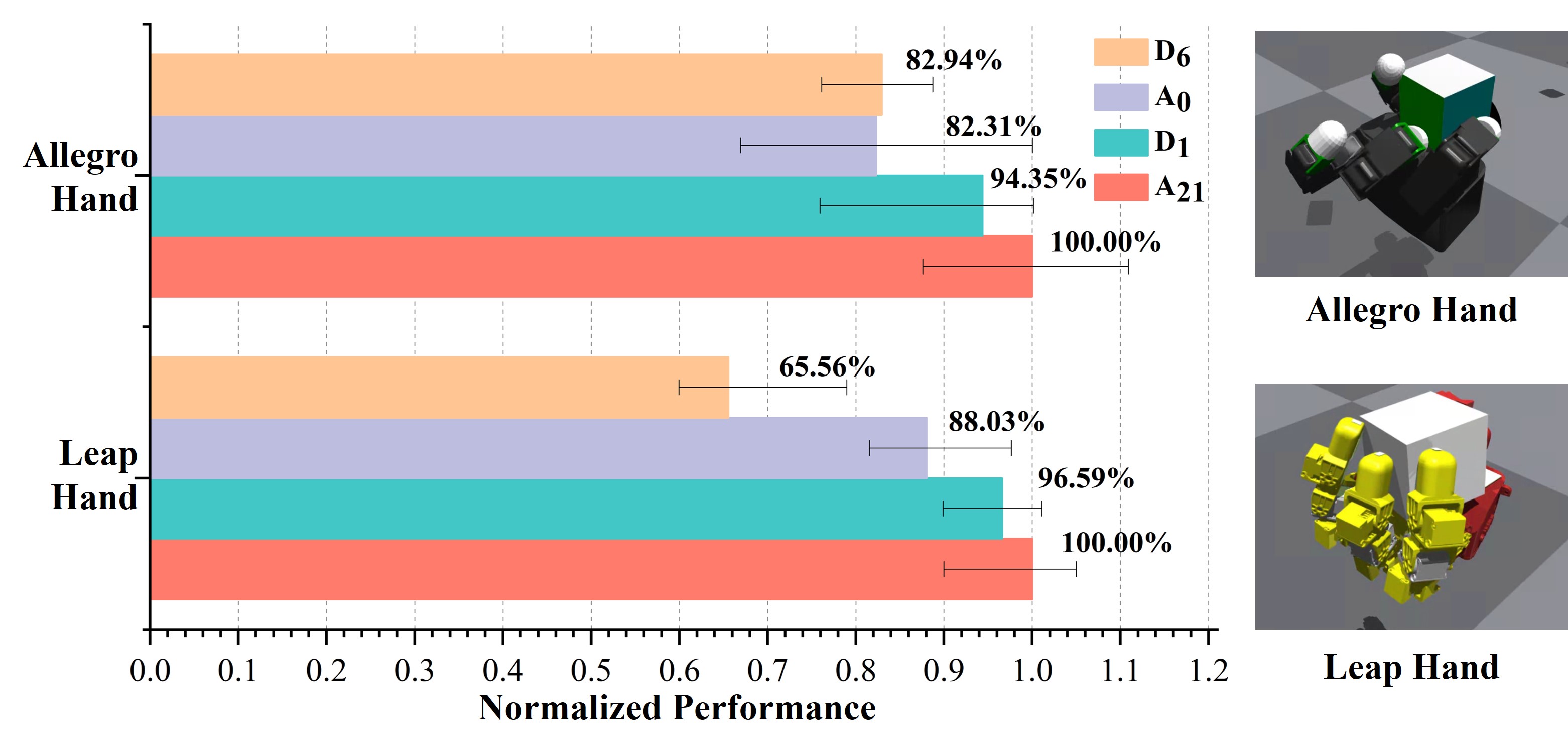}
\vspace{-2mm}
\caption{\footnotesize Results of cross-robot verification on Allegro Hand and Leap Hand. The bar chart illustrates the normalized performance across four configurations: $A_{21}$, $A_0$, $D_1$, and $D_6$. Error bars represent the standard deviation. All results are normalized relative to the $A_{21}$ configuration.}
\vspace{-2mm}
\label{fig:cross_robot}
\end{figure}

\noindent Fig.~\ref{fig:cross_robot} shows results on the Allegro Hand and Leap Hand. The $D_1$ configuration retains approximately 95\% of $A_{21}$ performance on both platforms (94.35\% for Allegro, 96.59\% for Leap). In contrast, $D_6$ performs at or below the sensor-free $A_0$ on both hands (82.94\% vs.\ 82.31\% for Allegro; 65.56\% vs.\ 88.03\% for Leap).

Two observations merit discussion. First, the Leap Hand shows a larger gap between $D_1$ and $D_6$ (31 percentage points) than the Allegro Hand (11 percentage points). This difference likely stems from the Leap Hand's more anthropomorphic finger proportions, which more closely replicate the Shadow Hand's kinematic structure and thus more faithfully inherit both the beneficial and detrimental sensor effects identified in our analysis. Second, on both platforms, the relative ordering $A_{21} \geq D_1 > A_0 > D_6$ is preserved---the same ordering observed on the five-fingered Shadow Hand. This consistency across platforms with different actuator types (tendon-driven vs.\ direct-drive), joint counts (16 vs.\ 20 DoF), and finger numbers (4 vs.\ 5) suggests that the identified importance rankings reflect fundamental contact patterns of in-hand manipulation rather than platform-specific kinematic artifacts. The practical implication is that practitioners designing sensor layouts for new hand platforms can use our Shadow Hand-derived rankings as an informed starting point, substantially narrowing the configuration search space.

\begin{figure}[!t]
\centering
\includegraphics[width=\columnwidth]{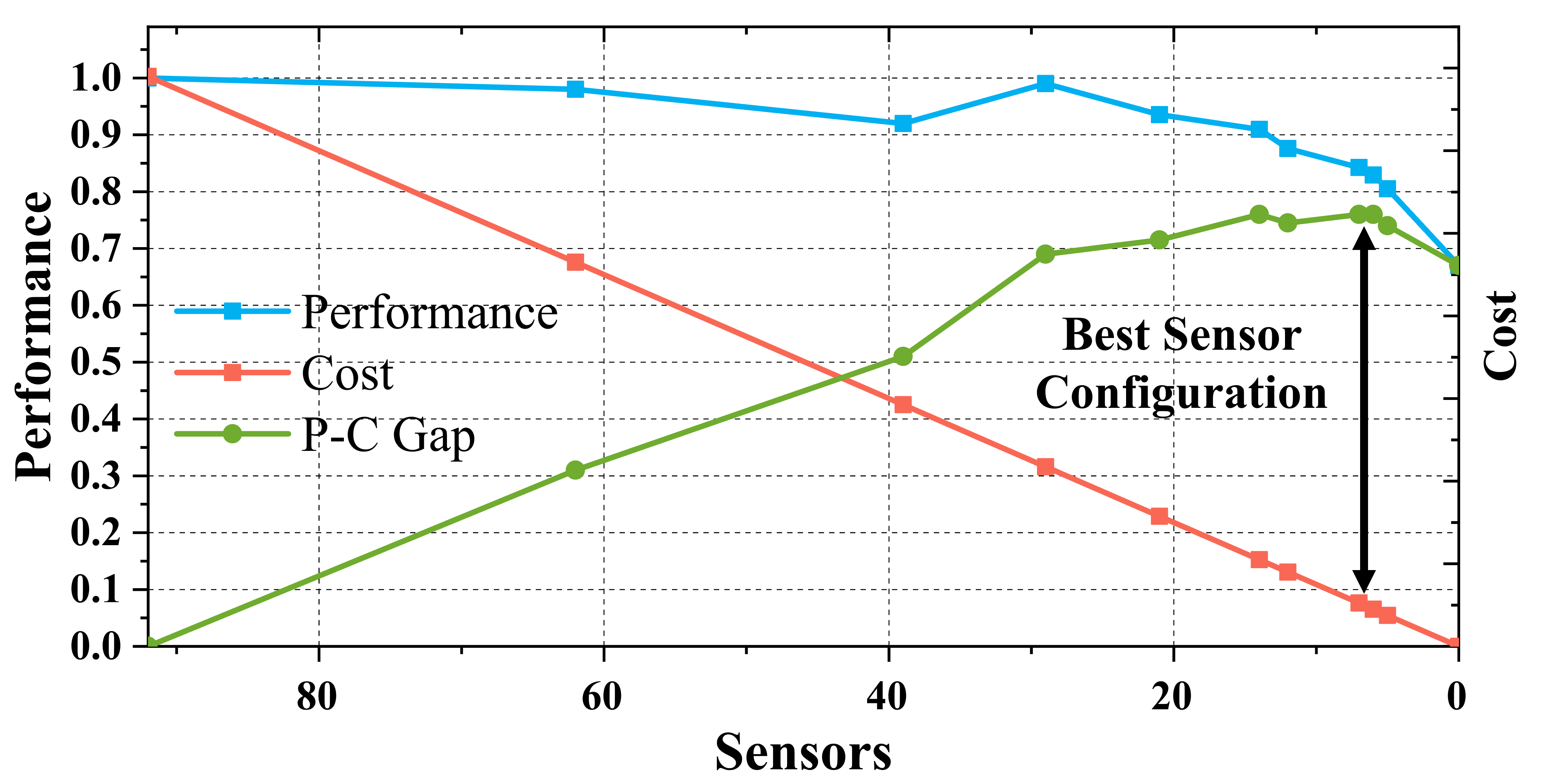}
\caption{\footnotesize Trends of average performance and cost for five tasks under different numbers of sensors. The green line represents the gap between performance and cost. A higher gap represents higher cost-effectiveness.}
\label{fig:cost_gap}
\vspace{-6mm}
\end{figure}

\subsection{Deployment Guidelines}

\noindent Based on the quantitative sensor importance analysis, we propose two distinct tactile deployment strategies tailored to different operational constraints (Fig.~\ref{fig:cost_gap}):

\textit{Performance-Oriented Strategy:} The $D_1$ configuration (14 sensors) retains all positively-contributing sensors, achieving $\sim$90\% of $A_{92}$ performance with robust zero-shot transferability. Compared to $A_{21}$, $D_1$ removes 7 sensors (33\% reduction) with no statistically significant performance loss on any of the five tasks (Table~\ref{tab:significance}), making it the recommended default configuration for general-purpose manipulation. For tighter budgets, sensors can be removed following the importance ranking down to $D_5$ (5 sensors), which maintains $>$80\% average performance at the cost of increased task-specific variance and reduced transfer robustness.

\textit{Cost-Effective Strategy:} To identify the optimal trade-off between sensing capability and system complexity, we evaluate the performance-cost gap (Fig.~\ref{fig:cost_gap}). The $D_3$ configuration emerges as the optimal point, yielding the maximum margin between performance and resource expenditure. This optimization principle remains extensible to alternative cost models by identifying the configuration that maximizes the divergence between the normalized cost function and the performance curve. Given that hardware complexity and data synchronization overhead often scale super-linearly with sensor count, prioritizing a minimal set of critical sensors significantly enhances the overall system cost-effectiveness.

%Analysis of the performance-cost gap (Fig.~\ref{fig:cost_gap}) identifies $D_3$ as the configuration maximizing the margin between normalized performance and normalized cost. Concretely, $D_3$ retains 82.1\% performance on the Block task while using only 9 sensors---a cost-effectiveness ratio of 9.1\% performance per sensor, compared to 4.8\% for $D_1$ and 1.0\% for $A_{92}$. Since hardware complexity typically grows super-linearly with sensor count due to wiring, calibration, and signal processing overhead, this gap analysis provides a principled basis for balancing performance against integration effort. For applications where a custom cost model applies (e.g., per-sensor maintenance costs, power budget constraints), practitioners can substitute their cost function into Fig.~\ref{fig:cost_gap} and identify the configuration that maximizes the revised gap.

% ============================================================
% VII. LIMITATIONS
% ============================================================
\section{Limitations}

\noindent Several limitations should be considered when interpreting our results. First, all experiments use binary (contact/no-contact) sensor outputs. While prior work~\cite{ref8} demonstrates comparable DRL performance between analog and binary tactile inputs in simulation, importance rankings may shift for continuous-valued sensors that capture force magnitude and direction. Second, our analysis is conducted entirely in simulation. Real-world factors, including sensor noise, drift, and contact model inaccuracies, could alter the relative importance of individual sensors. However, the sensor importance identified in this work is validated across two simulators and three robotic hand platforms, suggesting strong potential for practical deployment on real hardware. Third, although cross-task transfer results suggest reasonable generalization, the anomalous performance of $D_6$ on the Pentagonal Prism task indicates that sensor importance is partially task-dependent. For highly specialized applications, we recommend re-running the importance analysis on task-specific training data.

% ============================================================
% VIII. CONCLUSION
% ============================================================
\section{Conclusion}

\noindent This paper presented a two-stage framework for optimizing tactile sensor configurations in learning-based dexterous manipulation. Coarse empirical pruning reduces the Shadow Hand's 92-sensor array to 21 sensors with 93\% performance retention, and subsequent GPR-guided active learning with Lasso regression quantifies the importance of each remaining sensor. Our central finding is that sensor contributions are highly non-uniform: a 14-sensor subset concentrated on the thumb, ring finger, and little finger preserves over 90\% of full-array performance, while middle-finger sensors actively degrade policy learning. These rankings generalize across five manipulation tasks, three robot hands and two physics simulation platforms. The resulting deployment guidelines provide practitioners with explicit performance-cost trade-offs for designing scalable, cost-efficient tactile systems.

% ============================================================
% REFERENCES
% ============================================================
\bibliographystyle{IEEEtran}
\bibliography{references}

\end{document}